\documentclass{article}

\usepackage[preprint]{neurips_2026}

\usepackage[utf8]{inputenc} 
\usepackage[T1]{fontenc}    
\usepackage{hyperref}       
\usepackage{url}            
\usepackage{booktabs}       
\usepackage{amsfonts}       
\usepackage{nicefrac}       
\usepackage{xcolor}         
\usepackage{graphicx}
\usepackage{algorithm}
\usepackage{algorithmic}
\usepackage{amsthm}
\usepackage{multirow}
\usepackage{caption}

\usepackage{array}
\usepackage{pifont}
\newcommand{\cmark}{\textcolor{green!50!black}{\ding{51}}}
\newcommand{\xmark}{\textcolor{red!70!black}{\ding{55}}}
\newcolumntype{T}[1]{>{\raggedright\arraybackslash\vspace{0pt}}p{#1}}
\newcommand{\caseimg}[1]{%
  \vspace{0pt}\includegraphics[width=\linewidth]{case_studies/#1}%
}

\newcommand{\method}{GeoWorld-VLM}

\title{GeoWorld-VLM: Geometry from World Models for Vision-Language Models}

\author{%
  Renjie Gu$^{*}$, Kaichen Zhou$^{*}$, Yan Luo, Mengyu Wang \\
  Harvard AI and Robotics Lab \\ 
  Kempner Institute for the Study of Natural and Artificial Intelligence \\
  Harvard University \\
  $^{*}$Equal contribution as co-first authors
}

\begin{document}

\maketitle

\begin{center}
\includegraphics[width=\linewidth]{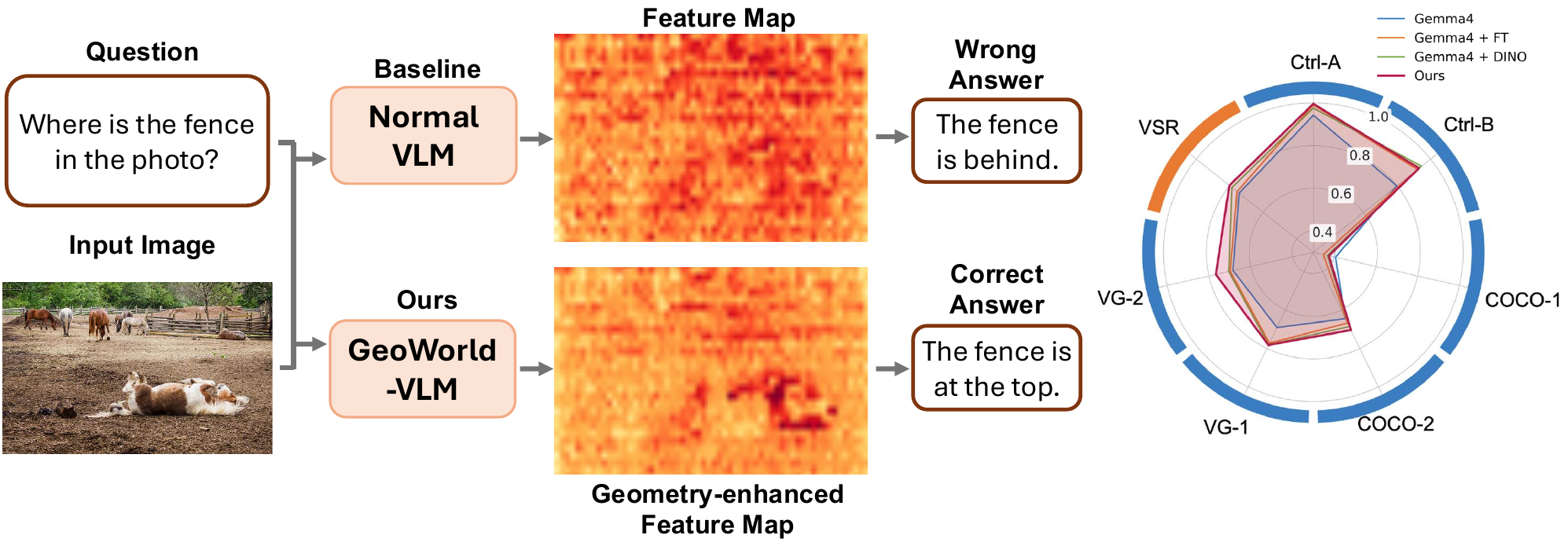}
\end{center}
\captionsetup{type=figure}
\captionof{figure}{%
  \textbf{Overview.} Given an input image and a spatial reasoning question, GeoWorld-VLM enhances the spatial understanding of standard vision-language models by injecting world-model priors at the feature-map level. Compared with the original VLM features, GeoWorld-VLM produces more geometry-aware representations, leading to clearer spatial grounding and improved answer accuracy. As shown on the right, our method consistently outperforms strong baselines, including the original Gemma-4, fine-tuned Gemma, and fine-tuned Gemma with DINO features, across diverse spatial reasoning benchmarks such as What’sUp and VSR.}
  \label{fig:tesser}
\vspace{0.3cm}

\begin{abstract}
Modern Vision-Language Models (VLMs) achieve strong semantic recognition, yet remain brittle on elementary spatial relations such as left of, on, behind, and between. 
One cause of this failure arises before language reasoning begins: the visual pathway may compress or discard critical 3D structural cues during feature extraction, so the language model receives image representations that are already insufficient for reliable spatial judgment.
We introduce GeoWorld-VLM, a VLM-side distillation framework that transfers geometric structure from frozen camera-conditioned video world models into VLMs. 
GeoWorld-VLM fine-tunes only the image encoder and multimodal projector, aligning post-projector image features with intermediate world-model representations while leaving the main backbone frozen. 
Given images, a prompt, and a sampled camera trajectory, the world-model teacher converts static visual input into a synthetic multi-view spatial signal.
Training combines spatial answer supervision, teacher-student feature alignment, and a preservation anchor to the original VLM. 
Since the language model remains frozen, GeoWorld-VLM preserves the original model’s linguistic capabilities while attributing spatial improvements to the enhanced visual pathway.
To evaluate the effectiveness and generality of the proposed method, we apply GeoWorld-VLM to two distinct VLM architectures and observe consistent improvements across both backbones. GeoWorld-VLM improves performance by approximately 4\% on both the What’sUp and VSR benchmarks, suggesting that world-model-guided visual alignment generalizes across model structures and spatial reasoning datasets. Code could be found: \href{https://github.com/Harvard-AI-and-Robotics-Lab/GeoWorld-VLM}{\nolinkurl{Link}}.
\end{abstract}

\section{Introduction}

Vision-Language Models (VLMs) have become fluent semantic recognizers, but they remain unreliable spatial reasoners~\cite{liu2025abstract,wu2025qwen}. Even strong contemporary VLMs struggle with elementary relations such as \emph{left of}, \emph{on top of}, \emph{behind}, and \emph{between}. These failures are surprising because the required reasoning is not linguistic: the model often knows the object names and understands the question, yet still fails to recover the geometric relation. This suggests that the bottleneck lies not in language reasoning alone, but in the visual representation delivered to the language model.

A modern VLM can be viewed as a composition of three modules~\cite{hurst2024gpt,luo2025visual,yang2025magma}:
\begin{equation}
\mathrm{VLM}(x_v,x_t)
=
\mathrm{LLM}\left(
\left[
P_\theta(V_\phi(x_v)), E(x_t)
\right]
\right),
\end{equation}
where $V_\phi$ is the vision encoder, $P_\theta$ is the multimodal projector, $E$ is the text encoder and the LLM performs language-conditioned reasoning. 
This decomposition exposes three possible places to repair spatial reasoning: the LLM, the text encoder, or the vision encoder \& projector. 
In this paper, we argue that improving the spatial understanding capability of the vision encoder and multimodal projector can effectively enhance VLM performance on spatial intelligence questions.

The key challenge is how to teach the visual pathway geometric structure without relying on real multi-view supervision. 
Recent world models have demonstrated strong spatial understanding and prediction capabilities across diverse scenarios with sparse observation~\cite{wan2025wan,yang2024cogvideox,team2026advancing,zhou2026gem}. 
As shown in Figure~\ref{fig:tesser}, motivated by this observation, we introduce \textbf{GeoWorld-VLM}, a VLM-side distillation framework that transfers spatial structure from frozen camera-conditioned world models into VLMs.
Given images, a text prompt, and a sampled egocentric camera trajectory, the world-model teacher produces an intermediate representation that captures how the scene would evolve under viewpoint motion, offering motion-aware geometric cues that are difficult to recover from the static view alone.
These cues provide an implicit signal about object layout, relative position, and view-dependent structure, which are often compressed or weakened in standard VLM visual representations.
GeoWorld-VLM distills this signal into the VLM by aligning its post-projector image features with the teacher representation in a shared feature space, thereby encouraging the visual pathway to preserve motion-aware spatial structure and scene geometry before these visual tokens are passed to the language model for reasoning.
Crucially, GeoWorld-VLM leaves the language model frozen. 
Within the VLM, we update only the image encoder and multimodal projector, together with lightweight training-time alignment heads. Training combines spatial answer supervision, teacher-student feature alignment, and a preservation anchor to a frozen copy of the original VLM. 
This design isolates spatial improvement to the visual interface: text-only behavior is preserved by construction, while the visual tokens entering the LLM become more geometry-aware.

To evaluate the effectiveness and generality of the proposed method, we instantiate our plug-in framework with two representative VLM backbones, Gemma~\cite{team2024gemma} and InternVL3.5~\cite{wang2025internvl3}, and evaluate them on the What’sUp and VSR benchmarks. Across both architectures and datasets, our method achieves consistent improvements under world-model supervision. Ablation studies further show that supervision from camera-conditioned world models is more effective than using conventional vision foundation models such as DINO~\cite{caron2021emerging} or geometry foundation models such as VGGT~\cite{wang2025vggt}.

\begin{itemize}
    \item \textbf{World-model distillation for VLM spatial reasoning.}
    We introduce GeoWorld-VLM, a framework that uses frozen camera-conditioned video world models as geometry teachers for VLMs. By sampling camera trajectories from input images, the teacher provides synthetic multi-view supervision without requiring real multi-view data.

    \item \textbf{A vision-side repair that preserves the language model.}
    GeoWorld-VLM fine-tunes only the vision encoder and multimodal projector while keeping the LLM and language encoder frozen. This isolates the spatial intervention to the visual interface and preserves text-only behavior by construction.

    \item \textbf{Broad validation across spatial and embodied benchmarks.}
    We evaluate GeoWorld-VLM across multiple spatial reasoning benchmarks and VLM backbones, demonstrating consistent improvements over the base models and stronger spatial understanding.
\end{itemize}
\section{Related Work}

\subsection{Spatial and 3D Reasoning in Vision-Language Models}


Modern vision-language models such as GPT-4o~\citep{hurst2024gpt4o}, GPT-5~\citep{openai2025gpt5}, Gemini~\citep{team2024gemini15}, Claude~\citep{anthropic2024claude3}, Qwen-VL~\citep{bai2023qwenvl}, Qwen2.5-VL~\citep{bai2025qwen25vl}, Qwen3-VL~\citep{qwen2025qwen3vl}, LLaVA~\citep{liu2023llava}, BLIP-2~\citep{li2023blip2}, InstructBLIP~\citep{dai2023instructblip}, PaLI~\citep{chen2022pali}, and PaliGemma~\citep{beyer2024paligemma} have shown strong semantic visual understanding, but they remain limited in spatial and 3D reasoning. Benchmarks such as SAT~\citep{ray2024sat}, What's\,Up~\citep{kamath2023whatsup}, BLINK~\citep{fu2024blink}, EmbSpatial-Bench~\citep{du2024embspatial}, and PhysBench~\citep{chow2025physbench} reveal persistent failures on relative position, relative depth, perspective taking, ego-motion, object movement, and physical consequence prediction. These failures suggest that VLMs can often recognize objects and parse questions, yet still lack the geometric structure needed to reason about spatial relations.
Several works attempt to bridge this gap by injecting 3D representations into language or vision-language models. 
3D-LLM~\citep{hong20233d} introduces 3D scene features into language models for embodied reasoning. 
Cube-LLM~\citep{cho2024language} and Language-Image Models with 3D Understanding explore 3D-aware representations for multimodal reasoning. 
MiniGPT-3D~\citep{tang2024minigpt} aligns point clouds with language models using 2D priors, while ShapeLLM focuses on 3D object understanding through point-cloud encoders and 3D instruction tuning. 
3UR-LLM~\citep{xiong20253ur} further extends multimodal language models toward 3D scene understanding. 
These approaches show that explicit 3D information can improve spatial reasoning, but typically require 3D supervision, point-cloud inputs, specialized architectures, or additional training.
Training-free or test-time approaches provide a more flexible alternative. 
APC~\citep{lee2025perspective} uses abstract perspective changes to support perspective-aware reasoning. 
VAGEN~\citep{wang2025vagen} applies multi-turn reasoning and world-model interaction for spatial intelligence. 
SandboxVLM~\citep{liu2025abstract} constructs abstract 3D bounding boxes through proxy elevation, multi-view voting, and box rendering, showing that coarse structural representations~\cite{wang2025vggt, zhou2025page} can help VLMs reason spatially without retraining. 
Our work follows the same broad goal of improving VLM spatial intelligence, but differs in where the improvement is applied. Rather than providing external 3D context at test time, GeoWorld-VLM distills geometry into the VLM's own vision side, improving the visual tokens consumed by the frozen language model.

\subsection{World Models and Geometric Supervision for VLMs}

World models and video generative models provide a natural source of geometric supervision because they could predict how scenes evolve under time, camera motion, and object interaction~\citep{wan2025wan,team2026advancing,yang2024cogvideox}. 
Classical world-model approaches such as Dreamer~\citep{wang2024worlddreamer}, DreamerV2~\citep{zhao2025drivedreamer} and DreamerV3~\citep{hafner2023mastering} learn predictive latent dynamics for control and planning. 
More recent generative world models and video models, including Genie~\citep{bruce2024genie}, VideoPoet~\citep{kondratyuk2023videopoet}, Sora-style video models~\citep{lin2024open}, CogVideoX~\citep{yang2024cogvideox}, Wan~\citep{wan2025wan}, HunyuanVideo~\citep{kong2024hunyuanvideo}, Cosmos~\citep{agarwal2025cosmos}, and Stable Virtual Camera~\citep{zhou2025stable}, scale this idea to rich visual scenes and viewpoint-conditioned generation. 
These models implicitly encode cues such as parallax, occlusion, support, relative depth, and object persistence.
Recent VLM reasoning methods have begun to exploit such models. 
MindJourney~\citep{yang2025mindjourney} uses imagined visual trajectories at test time to help VLMs reason about spatial questions. 
These works suggest that world models contain useful spatial priors, but most use them externally: they generate images, videos, or reconstructed 3D contexts that are then fed back to a VLM during inference.
GeoWorld-VLM instead uses the world model as a feature teacher. The teacher receives the initial image, a text condition, and a sampled camera trajectory, producing an intermediate representation shaped by camera-conditioned prediction. 
We distill this representation into the VLM's post-projector image feature, while keeping the LLM frozen. 
This differs from static feature distillation methods based on DINO~\cite{caron2021emerging}, DINOv2~\cite{oquab2023dinov2}, CLIP~\citep{radford2021learning}, or SigLIP~\citep{zhai2023sigmoid}, which provide strong semantic representations but do not explicitly encode viewpoint-dependent geometry. 
It also differs from world-model-only image conditioning, where the teacher receives no camera trajectory and therefore behaves more like a static image encoder. 
By conditioning the world-model teacher on both camera motion and textual context, GeoWorld-VLM transforms a single static image into a synthetic multi-view supervision signal that encodes how the scene would change under viewpoint variation, and transfers this geometry-aware information into the VLM's visual interface through feature-level alignment.

\section{Method}
\label{sec:method}

\begin{figure*}[t]
\begin{center}
\includegraphics[width=\linewidth]{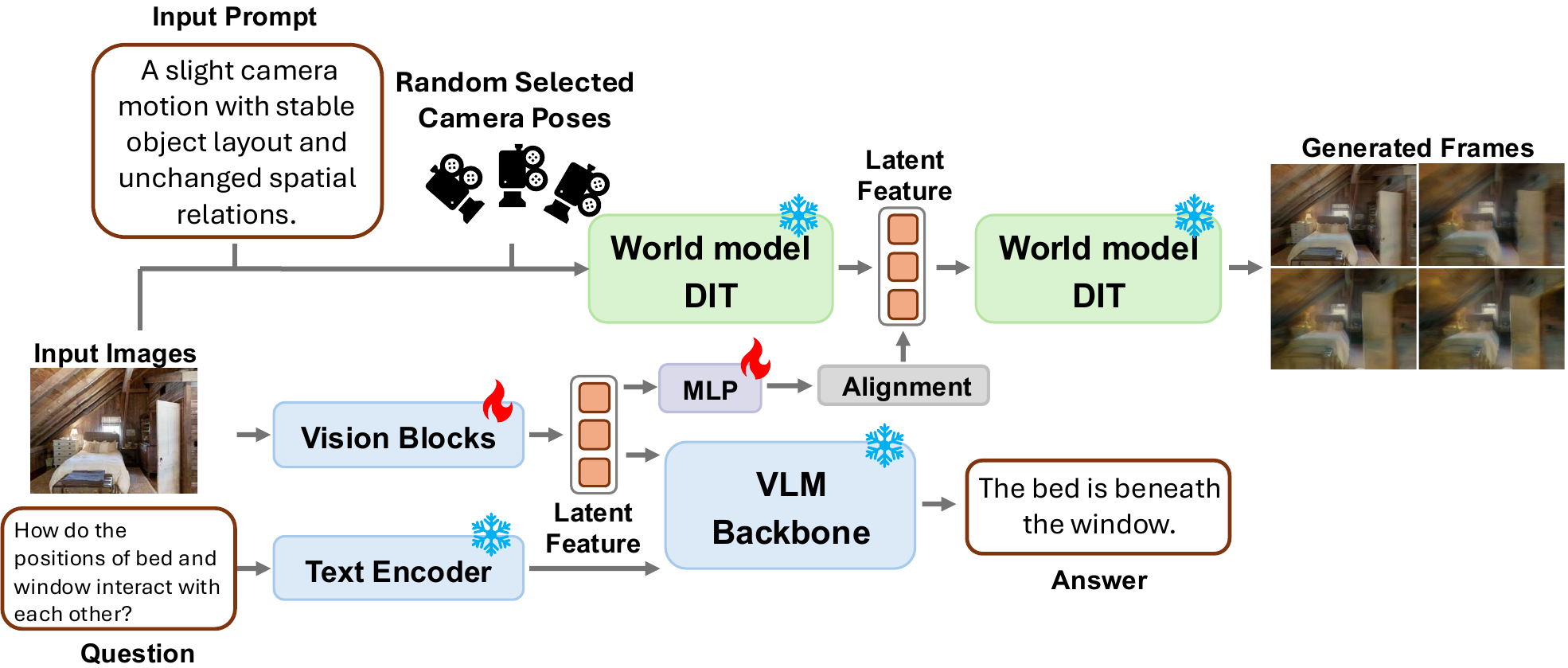}
\end{center}
\caption{\textbf{\method{}.} During training, \method{} fine-tunes only the vision blocks including vision encoder and multimodal projector. It aligns the latent features produced by the VLM vision encoder with intermediate world-model representations, where the world model takes the input image, text prompt, and randomly sampled camera poses as input. At inference time, \method{} no longer requires the world model and can perform standard VLM inference directly.}
\label{fig:pipeline}
\end{figure*}

We introduce \textbf{GeoWorld-VLM}, a vision-side distillation method that transfers geometric structure from a frozen camera-conditioned video world model into a VLM~\cite{beyer2024paligemma, wang2025internvl3, team2026qwen3}. 
As shown in Figure~\ref{fig:pipeline}, our design follows from a simple hypothesis: spatial failures arise because the visual tokens entering the LLM lack sufficient 3D structure. 
We therefore keep the LLM frozen and update only the image encoder, multimodal projector, and lightweight alignment heads.

\subsection{Vision-side formulation}
Following most VLM architectures, we design a VLM:
\begin{equation}
\mathrm{VLM}(x_v,x_t)
=
\Psi\left([P_\theta(V_\phi(x_v)),E(x_t)]\right),
\end{equation}
where $V_\phi$ is the image encoder, $P_\theta$ is the multimodal projector, $E$ is the text encoder and $\Psi$ is the main LLM backbone. GeoWorld-VLM updates only $(\phi,\theta)$. This isolates the intervention to the representation consumed by the LLM: text-only behavior is unchanged by construction, and any spatial improvement must come from better visual tokens rather than language-side adaptation.

\subsection{Camera-conditioned world-model teacher}
Static image teachers such as DINO~\cite{caron2021emerging,oquab2023dinov2,simeoni2025dinov3} provide strong semantic features, but they do not explicitly encode how a scene changes under viewpoint motion. This distinction matters for spatial reasoning: relations such as behind, between, support, and occlusion are revealed not only by what objects are present, but by how their projections change as the camera moves.
For each training image $x_v$, we sample an egocentric camera trajectory:
\begin{equation}
\pi=(\pi^{(1)},\ldots,\pi^{(F)}),
\end{equation}
from translations and rotations, including forward, backward, left, right, forward-left, backward-left, forward-right, backward-right, and other movements, as provided in LingBot-World~\cite{team2026advancing}. The frozen world-model teacher receives the initial images, the camera trajectory, and a text condition: $(x_v,\pi,c)$.
Here, $c$ is "A slight camera motion with stable object layout and unchanged spatial relations.". The teacher predicts the scene under the sampled camera motion, and we extract an intermediate hidden representation $g(x_v) = \mathcal{T}_{b^\star}(x_v,\pi,c, z_{t^\star})$ where $z_{t^\star}$ is the noised video latent at timestep $t^\star$ and $\mathcal{T}_{b^\star}$ denotes the hidden state at teacher block $b^\star$.
The camera trajectory is essential. If the teacher receives only images, its representation collapses toward a static semantic embedding. Conditioning on camera motion instead forces the teacher feature to encode counterfactual views, making a single image behave like a synthetic multi-view training signal, which could be helpful for spatial understanding.

\subsection{World-to-VLM alignment}

The student produces post-projector image features $h_{\phi,\theta}(x_v)=P_\theta(V_\phi(x_v))$.
Since the VLM and world model have different representation spaces, we use two lightweight projection heads:
\begin{equation}
u_s=f^s_\psi(h_{\phi,\theta}(x_v)),
\qquad
u_t=f^t_\psi(g(x_v)).
\end{equation}
The alignment loss is
\begin{equation}
\mathcal{L}_{\mathrm{align}}
=
1-
\left\langle
\frac{u_s}{\|u_s\|_2},
\frac{u_t}{\|u_t\|_2}
\right\rangle.
\end{equation}
We align at the post-projector level because this representation is directly consumed by the frozen LLM for multimodal reasoning. Aligning earlier visual features may improve the encoder but does not guarantee that geometry survives the multimodal interface.

\subsection{Preserving the original interface}

World-model alignment alone can move the visual tokens away from the distribution expected by the frozen LLM. To control this shift, we keep a frozen copy of the original vision side, $(V_{\phi'},P_{\theta'})$, and add a post-projector preservation loss:
\begin{equation}
\mathcal{L}_{\mathrm{preserve}}
=
\left\|
\frac{h_{\phi,\theta}}{\| h_{\phi,\theta}\|_2}
-
\frac{h_{\phi',\theta'}}{\| h_{\phi',\theta'}\|_2}
\right\|_2^2.
\end{equation}
The alignment and preservation losses play complementary roles: alignment injects geometry from the world model, while preservation keeps visual tokens compatible with the frozen language stack.
\subsection{Training objective}

GeoWorld-VLM combines spatial supervision, world-model alignment, and interface preservation:
\begin{equation}
\mathcal{L}
=
\mathcal{L}_{\mathrm{task}}
+
\lambda_{\mathrm{align}}\mathcal{L}_{\mathrm{align}}
+
\lambda_{\mathrm{preserve}}\mathcal{L}_{\mathrm{preserve}}.
\end{equation}
Here, $\mathcal{L}_{\mathrm{task}}$ is the option-restricted cross-entropy loss for spatial multiple-choice supervision. Gradients update only $(\phi,\theta,\psi)$; the LLM, the world-model teacher, and the frozen reference VLM remain fixed.
This design separates the source of supervision from the locus of adaptation. The world model supplies geometry through camera-conditioned prediction; the VLM absorbs that structure only through its vision side; and the LLM remains unchanged. As a result, GeoWorld-VLM tests a focused claim: many spatial reasoning failures in VLMs can be repaired by improving the visual interface itself rather than retraining or modifying the language model.
\section{Experiments}
\label{sec:experiments}
\begin{table*}[t]
\centering
\small
\setlength{\tabcolsep}{4.5pt}
\resizebox{\textwidth}{!}{
\begin{tabular}{lcccccccc}
\toprule
\textbf{Model}
& \textbf{Ctrl-A}
& \textbf{Ctrl-B}
& \textbf{COCO-1}
& \textbf{COCO-2}
& \textbf{VG-1}
& \textbf{VG-2}
& \textbf{VSR}
& \textbf{Overall} \\
\midrule
Gemma4
& 94.17 & 79.90 & \textbf{40.57} & 64.55 & 69.14 & 68.49 & 74.36 & 61.70 \\
Gemma4 + FT
& \underline{99.03} & 92.16 & 34.79 & 66.82 & 77.24 & 69.86 & 75.56 & 62.71 \\
Gemma4 + DINO
& 97.57 & \textbf{94.61} & 36.74 & \underline{68.64} & \underline{78.10} & \underline{70.55} & \underline{78.63} & \underline{64.40} \\
Gemma4 + Ours
& \textbf{99.51} & \underline{93.14} & \underline{37.46} & \textbf{70.45} & \textbf{78.28} & \textbf{76.71} & \textbf{80.17} & \textbf{65.45} \\
\midrule
InternVL3.5-2B
& 64.08 & 81.37 & \textbf{40.12} & \underline{60.00} & \underline{65.69} & \underline{41.78} & 72.82 & 57.06 \\
InternVL3.5-2B + FT
& \underline{98.54} & \underline{96.08} & 31.76 & 55.00 & 61.38 & 41.10 & \textbf{83.59} & \underline{58.14} \\
InternVL3.5-2B + DINO
& 97.57 & \textbf{99.51} & 29.54 & 52.73 & 51.38 & 29.45 & \underline{83.25} & 54.81 \\
InternVL3.5-2B + Ours
& \textbf{99.51} & \textbf{99.51} & \underline{35.77} & \textbf{60.91} & \textbf{70.52} & \textbf{45.89} & 80.34 & \textbf{61.66} \\
\bottomrule
\end{tabular}
}
\caption{
Main results on the combined What'sUp+VSR evaluation suite.
GeoWorld-VLM consistently improves the overall score over the original model, task-only fine-tuning, and DINO-based distillation.
Best results are shown in bold, and second-best results are underlined.
}
\label{tab:main_wv_results}
\end{table*}

\subsection{Experimental Setup}
\label{subsec:exp_setup}

\paragraph{Baselines \& VLMs.}
We evaluate GeoWorld-VLM on two representative vision-language model backbones: \texttt{google/gemma-4-E4B-it} and \texttt{OpenGVLab/InternVL3\_5-2B-Instruct}~\cite{wang2025internvl35advancingopensourcemultimodal}.
For each VLM, we compare four variants.
\textit{Base} denotes the original VLM without additional spatial adaptation.
\textit{FT} denotes task-only fine-tuning, which removes the teacher-student feature alignment loss and the preservation anchor from GeoWorld-VLM and optimizes only the downstream spatial answer supervision.
This baseline tests whether the gains can be explained by target-task supervision alone.
\textit{DINO}~\citep{simeoni2025dinov3} denotes a DINOv3-based static feature distillation baseline, where the world-model teacher is replaced with a frozen DINOv3 encoder and the VLM visual representation is aligned with DINOv3 features.
This baseline tests whether generic visual representation distillation is sufficient for spatial reasoning.
\textit{Ours} denotes GeoWorld-VLM, which aligns post-projector image features with intermediate camera-conditioned world-model representations, allowing the student VLM to absorb motion-aware geometric cues from the frozen teacher while preserving compatibility with the original visual interface and leaving the language backbone unchanged.

\paragraph{Datasets.}
We evaluate GeoWorld-VLM on three spatial reasoning benchmarks: What'sUp~\citep{kamath2023whatsup}, Visual Spatial Reasoning (VSR)~\citep{liu2023visual}, and EmbSpatial-Bench~\citep{du2024embspatial}.
Following the evaluation protocol of prior spatial reasoning work~\citep{chen2025spatialreasoninghardvlms}, we combine What'sUp and VSR into a unified What'sUp+VSR suite.
This suite contains controlled spatial relation recognition, natural-image spatial QA, and image-text spatial verification.
Specifically, it covers object-centric relation recognition in controlled tabletop scenes, absolute and relative spatial localization in COCO and Visual Genome images, and binary verification of diverse spatial predicates in natural images.
We use 3,089 examples for training and the remaining 3,091 examples for evaluation.
For EmbSpatial, we use 2,000 examples for training and the remaining 1,640 examples for evaluation.
For What'sUp+VSR, we report accuracy on Controlled Images A/B, COCO-based QA, VG-based QA, and VSR, together with the overall average~\citep{lin2015microsoftcococommonobjects,zou2024vgbenchevaluatinglargelanguage}.
For EmbSpatial, we report relation-wise accuracy on six spatial relations: \textit{left}, \textit{right}, \textit{above}, \textit{under}, \textit{close}, and \textit{far}, together with the overall average.

\paragraph{Implementation Details.}
GeoWorld-VLM uses LingBot-World-Fast~\citep{robbyantteam2026advancingopensourceworldmodels} as a frozen camera-conditioned image-to-video teacher, extracting intermediate DiT features as geometry-aware supervision.
For each input image, the teacher takes a fixed lightweight prompt, ``A slight camera motion with stable object layout and unchanged spatial relations.'', and a randomly sampled egocentric camera trajectory.
On the student side, we align post-projector image features, which are directly consumed by the frozen language model.
Following Section~\ref{sec:method}, we freeze the language model and update only the image encoder, multimodal projector, and lightweight alignment heads.
All main fine-tuning experiments are repeated using three different random seeds, and we report the averaged results across these independent runs.
Unless otherwise specified, teacher features are extracted from the 24th teacher block using two denoising steps.
All methods are trained for three epochs under the same downstream spatial supervision, with detailed hyperparameters provided in Appendix~\ref{app:training_details}.

\begin{figure*}[t]
\begin{center}
\includegraphics[width=\linewidth]{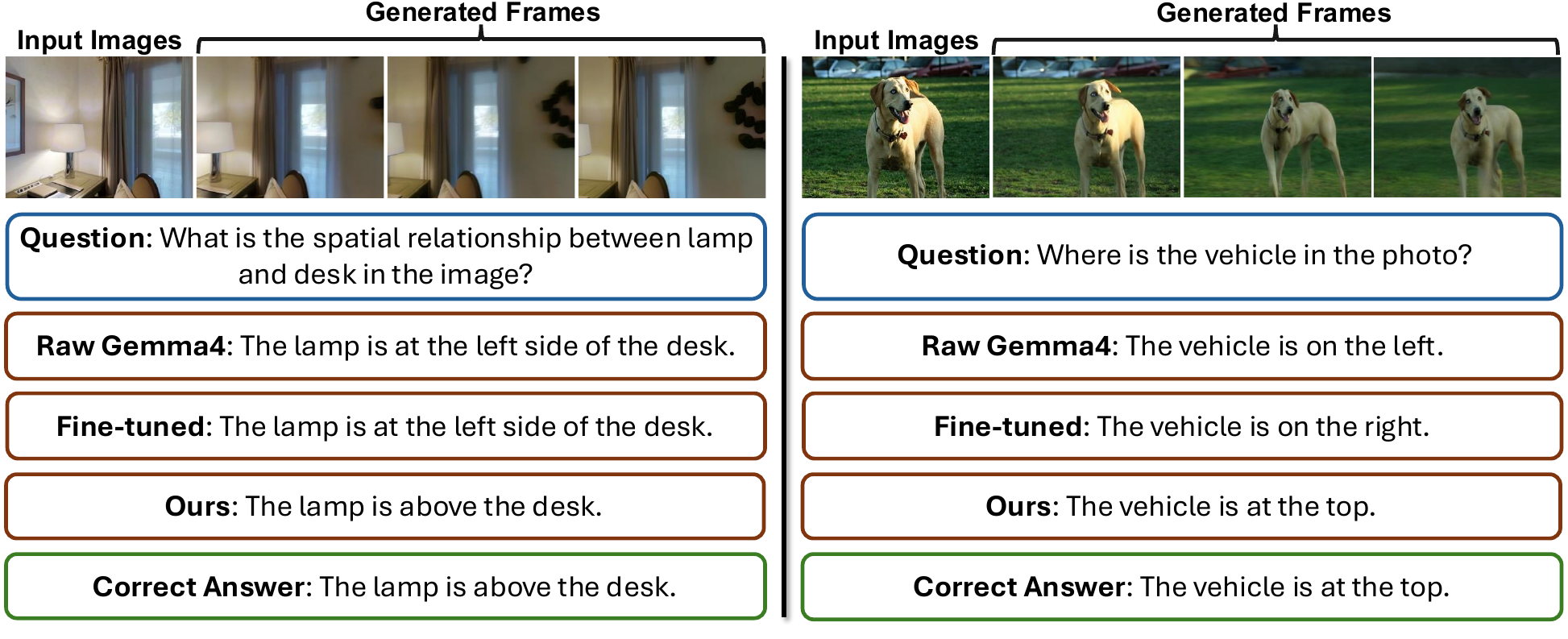}
\end{center}
\caption{
\textbf{Qualitative comparison.}
We compare the predictions of the original Gemma4, task-only fine-tuned Gemma4, and GeoWorld-VLM on representative spatial reasoning examples.
GeoWorld-VLM produces more spatially consistent answers than the baselines, illustrating the benefit of injecting camera-conditioned world-model supervision into the VLM visual pathway.
}
\label{fig:qualitative_examples}
\end{figure*}

\subsection{Main Results on What'sUp and VSR}
\label{subsec:main_results_wv}

Table~\ref{tab:main_wv_results} reports the main results on the What'sUp+VSR suite.
Across both VLM backbones, GeoWorld-VLM achieves the best overall performance.
For Gemma4, GeoWorld-VLM improves the overall score from $61.70$ to $65.45$, outperforming both task-only fine-tuning and DINOv3-based static feature distillation.
The gains are especially clear on natural-image spatial reasoning subsets such as COCO-2 and VG-2, where GeoWorld-VLM reaches $70.45$ and $76.71$, respectively.
This suggests that world-model supervision does not merely improve controlled relation recognition, but also transfers to more complex natural-image spatial reasoning.
The qualitative examples in Figure~\ref{fig:qualitative_examples} further show that GeoWorld-VLM produces more spatially grounded predictions than the baselines.

\begin{figure*}[t]
\begin{center}
\includegraphics[width=\linewidth]{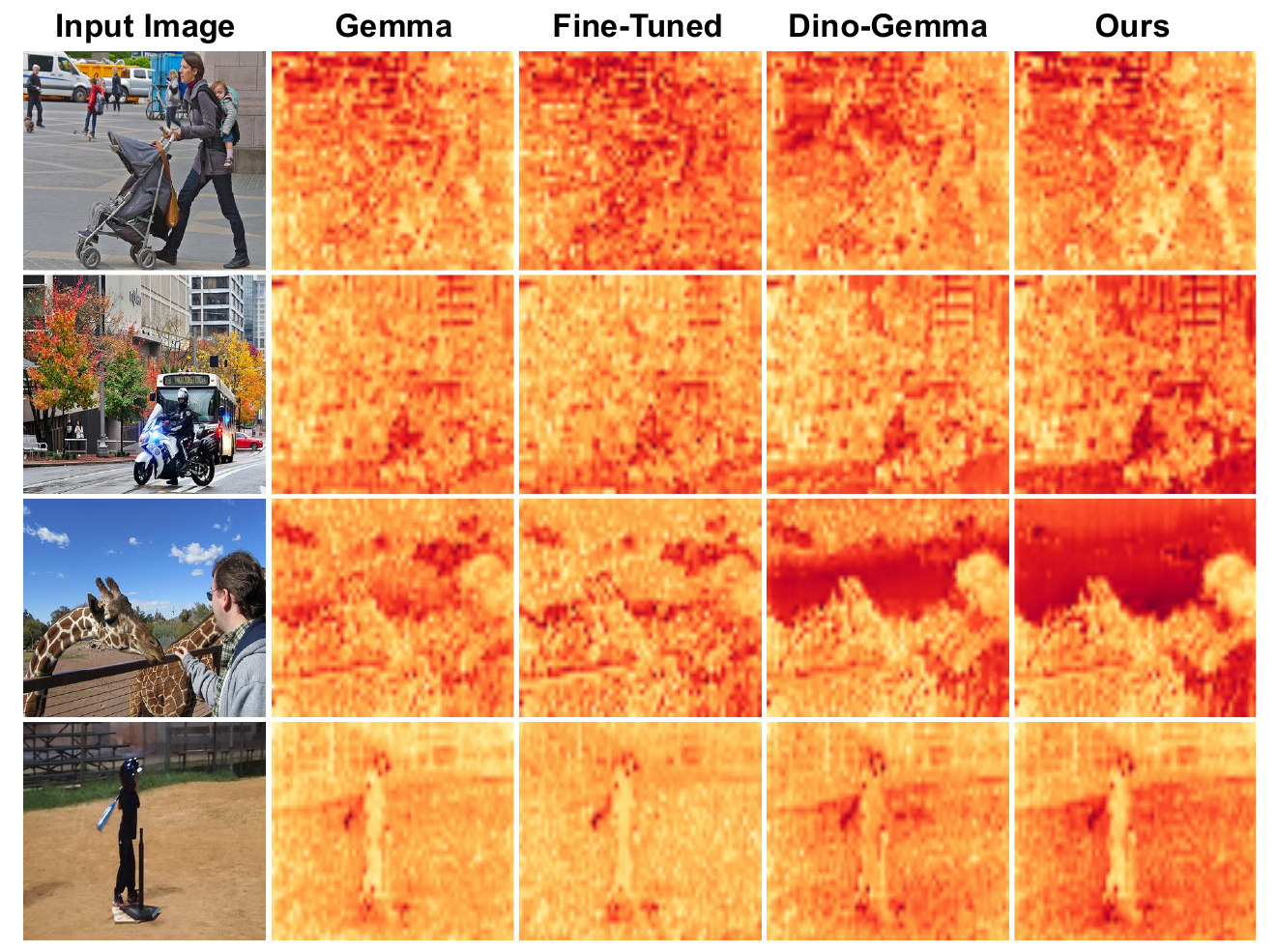}
\end{center}
\caption{
\textbf{Overall comparison on the What'sUp+VSR suite.}
GeoWorld-VLM improves spatial reasoning performance across diverse sub-benchmarks, showing consistent gains over the original VLM, task-only fine-tuning, and DINOv3-based static feature distillation.
}
\label{fig:wv_visualization}
\end{figure*}

For InternVL3.5-2B, GeoWorld-VLM improves the overall score from $57.06$ to $61.66$, yielding a $4.60$-point gain over the base model and a $3.52$-point gain over task-only fine-tuning.
In contrast, DINOv3-based feature distillation decreases the overall score to $54.81$.
This comparison is important because both DINO and GeoWorld-VLM introduce auxiliary feature-level supervision, but only GeoWorld-VLM provides camera-conditioned world-model representations.
The result indicates that static visual feature distillation may be insufficient, or even harmful, when the target task requires spatial judgment rather than generic visual discrimination.
By contrast, world-model features provide a more geometry-aware supervision signal, while the preservation anchor helps maintain compatibility with the frozen language model. We also observe that the gains are not uniform across all subsets.
For example, COCO-1 remains challenging for all adapted variants, and task-only fine-tuning can even reduce performance on this subset.
This suggests that spatial adaptation may shift the visual interface toward the training distribution if it is not properly regularized.
GeoWorld-VLM alleviates this issue by combining world-model supervision with the preservation anchor, improving the overall score while maintaining competitive performance across individual subsets. Figure~\ref{fig:wv_visualization} visualizes the overall comparison on the What'sUp+VSR suite.
The performance pattern is consistent with Table~\ref{tab:main_wv_results}: GeoWorld-VLM improves the visual pathway with geometry-aware supervision and achieves stronger spatial grounding without modifying the frozen language model.

\subsection{Relation-wise Results on EmbSpatial}
\label{subsec:embspatial_results}

\begin{table*}[t]
\centering
\small
\setlength{\tabcolsep}{6pt}
\resizebox{\textwidth}{!}{
\begin{tabular}{lccccccc}
\toprule
\textbf{Model}
& \textbf{Left}
& \textbf{Right}
& \textbf{Above}
& \textbf{Under}
& \textbf{Close}
& \textbf{Far}
& \textbf{Overall} \\
\midrule
Gemma4
& 84.53 & 60.57 & 60.45 & 55.35 & 53.99 & 47.76 & 60.55 \\
Gemma4 + FT
& \textbf{90.29} & \underline{77.78} & 86.57 & \underline{85.24} & 63.04 & \underline{58.96} & \underline{77.01} \\
Gemma4 + DINO
& 87.05 & 71.68 & \underline{87.69} & 84.50 & \underline{65.58} & \underline{58.96} & 75.91 \\
Gemma4 + Ours
& \underline{89.21} & \textbf{78.14} & \textbf{88.43} & \textbf{86.72} & \textbf{67.03} & \textbf{63.06} & \textbf{78.78} \\
\midrule
InternVL3.5-2B
& 78.42 & 58.42 & 64.55 & 28.78 & 43.12 & 42.16 & 52.68 \\
InternVL3.5-2B + FT
& \underline{86.69} & \textbf{82.80} & 82.84 & 82.29 & 65.58 & \textbf{68.66} & 78.17 \\
InternVL3.5-2B + DINO
& \textbf{87.41} & 78.85 & \textbf{90.67} & \underline{84.50} & \underline{67.75} & \underline{65.30} & \underline{79.09} \\
InternVL3.5-2B + Ours
& 84.89 & \underline{81.36} & \underline{89.18} & \textbf{87.82} & \textbf{69.93} & 64.93 & \textbf{79.70} \\
\bottomrule
\end{tabular}
}
\caption{
Relation-wise results on EmbSpatial.
GeoWorld-VLM improves the overall relation accuracy for both VLM backbones and shows strong gains on geometry-sensitive relations such as \textit{under}, \textit{close}, and \textit{far}.
The best result is shown in bold and the second-best result is underlined.
}
\label{tab:embspatial_results}
\end{table*}

Table~\ref{tab:embspatial_results} further evaluates GeoWorld-VLM on EmbSpatial.
This benchmark allows us to inspect whether the improvement comes from broad answer-pattern fitting or from better handling of specific spatial relations.
For Gemma4, GeoWorld-VLM improves the overall score from $60.55$ to $78.78$, outperforming task-only fine-tuning and DINOv3-based static feature distillation.
The gains are particularly clear on \textit{above}, \textit{under}, \textit{close}, and \textit{far}, which often require relative layout, support, depth, or distance estimation rather than object recognition alone.

For InternVL3.5-2B, GeoWorld-VLM improves the overall score from $52.68$ to $79.70$ and slightly surpasses DINOv3-based feature distillation.
The strongest gains appear on \textit{under} and \textit{close}, suggesting that the world-model teacher is especially useful for relations that depend on scene geometry and viewpoint.
Overall, the EmbSpatial results show that GeoWorld-VLM does not simply increase average accuracy; it strengthens spatial judgments that require geometry-aware visual representations. The relation-wise pattern also explains why the improvement is not identical across all categories.
Relations such as \textit{left} and \textit{right} can often be inferred from 2D image layout, so static visual features or task-only fine-tuning may already provide strong cues.
In contrast, relations such as \textit{under}, \textit{close}, and \textit{far} are more sensitive to viewpoint, support, and relative depth.
These are precisely the cases where camera-conditioned world-model supervision is expected to be most useful.

\subsection{Module Selection}
\label{subsec:module_selection}

\begin{table*}[t]
\centering
\small
\setlength{\tabcolsep}{4.5pt}
\resizebox{\textwidth}{!}{
\begin{tabular}{lcccccccc}
\toprule
\textbf{Model}
& \textbf{Ctrl-A}
& \textbf{Ctrl-B}
& \textbf{COCO-1}
& \textbf{COCO-2}
& \textbf{VG-1}
& \textbf{VG-2}
& \textbf{VSR}
& \textbf{Overall} \\
\midrule
Gemma4
& 94.17 & 79.90 & \textbf{40.57} & 64.55 & 69.14 & 68.49 & 74.36 & 61.70 \\
Gemma4 + I2V
& 99.03 & \textbf{94.12} & 35.68 & \underline{68.18} & \underline{77.59} & 67.81 & \underline{79.66} & \underline{64.01} \\
Gemma4 + I2V + Camera
& 98.06 & \underline{93.14} & 33.81 & 66.36 & 77.07 & 66.44 & \textbf{81.03} & 63.16 \\
Gemma4 + I2V + Prompt
& 98.54 & \underline{93.14} & 33.54 & 65.45 & 76.38 & \underline{70.55} & 78.46 & 62.61 \\
Gemma4 + I2V + Prompt + Camera
& \textbf{99.51} & \underline{93.14} & \underline{37.46} & \textbf{70.45} & \textbf{78.28} & \textbf{76.71} & 80.17 & \textbf{65.45} \\
\bottomrule
\end{tabular}
}
\caption{
Module selection results on the What'sUp+VSR suite.
We compare different teacher-side conditioning modules, including image-to-video features alone and variants with prompt or camera conditioning.
Based on the overall performance, we select image-to-video features with both prompt and camera conditioning as the default configuration of GeoWorld-VLM.
The best result is shown in bold and the second-best result is underlined.
}
\label{tab:module_selection}
\end{table*}

Table~\ref{tab:module_selection} studies which teacher-side conditioning module provides the most effective supervision for GeoWorld-VLM.
Using image-to-video world-model features alone already improves the overall score from $61.70$ to $64.01$, showing that world-model representations contain spatial information beyond the original VLM visual tokens.
However, adding only camera conditioning or only prompt conditioning does not further improve the overall result.
The camera-only and prompt-only variants reach $63.16$ and $62.61$, respectively. This indicates that the two conditioning signals are complementary rather than independently sufficient.
Camera trajectories provide geometric transformation cues, but without relation-aware text conditioning, the teacher representation may not focus on the spatial relation queried by the downstream task.
Conversely, prompt conditioning alone identifies the relevant relation but lacks the viewpoint variation needed to expose geometry-aware structure.

The best performance is obtained when prompt and camera conditioning are used jointly.
This full configuration improves the overall score to $65.45$, outperforming the base model by $3.75$ points and the I2V-only teacher by $1.44$ points.
The improvement is especially clear on COCO-2 and VG-2, where the full configuration reaches $70.45$ and $76.71$, respectively.
These results suggest that the most useful teacher representation is both \emph{question-aware} and \emph{geometry-aware}: the prompt specifies the spatial relation of interest, while the camera trajectory encourages features that encode scene structure under viewpoint changes.
Therefore, we use I2V features with both prompt and camera conditioning as the default teacher configuration of GeoWorld-VLM.
Beyond module selection, we further conduct parameter ablations on world-model feature extraction layer, denoising step, and the alignment loss coefficient.
As shown in Appendix~\ref{app:ablation_details}, GeoWorld-VLM achieves the best overall performance when using the 24th world-model layer, two denoising steps, and $\lambda_{\mathrm{align}}=0.10$.
These results suggest that world-model supervision is most effective when the teacher features are sufficiently spatially structured but not overly dominated by generation-specific dynamics.

\subsection{Ablation Studies}
\label{subsec:ablation_studies}

After selecting the default teacher-side conditioning module, we further ablate two key design choices in GeoWorld-VLM: the world-model feature extraction layer and the denoising step used for teacher feature extraction.
Both factors directly affect the spatial signal transferred from the world-model teacher to the VLM visual interface.
Unless otherwise specified, all ablation studies are conducted using Gemma4 as the VLM backbone and are evaluated on the What'sUp+VSR suite under the same experimental setting.
More detailed analysis is provided in Appendix~\ref{app:ablation_details}.

\paragraph{World-model layer selection.}
Table~\ref{tab:layer_ablation} compares layers 23, 24, and 25 of the world-model teacher.
Layer 24 achieves the best overall performance, improving the base Gemma4 model from $61.70$ to $65.45$ and outperforming layers 23 and 25 by $2.38$ and $2.35$ points, respectively.
This suggests that the most transferable spatial supervision lies in an intermediate-to-late region of the world-model hierarchy, where the representation has incorporated camera- and prompt-conditioned spatial structure while remaining general enough to supervise the VLM visual pathway.

\begin{table*}[t]
\centering
\small
\setlength{\tabcolsep}{4.5pt}
\resizebox{\textwidth}{!}{
\begin{tabular}{lcccccccc}
\toprule
\textbf{Model}
& \textbf{Ctrl-A}
& \textbf{Ctrl-B}
& \textbf{COCO-1}
& \textbf{COCO-2}
& \textbf{VG-1}
& \textbf{VG-2}
& \textbf{VSR}
& \textbf{Overall} \\
\midrule
Gemma4
& 94.17 & 79.90 & \textbf{40.57} & 64.55 & 69.14 & 68.49 & 74.36 & 61.70 \\
Gemma4 + Ours, layer 23
& 97.57 & \underline{94.12} & 34.25 & 63.64 & \underline{77.07} & \underline{65.75} & \textbf{80.68} & 63.07 \\
Gemma4 + Ours, layer 24
& \textbf{99.51} & 93.14 & \underline{37.46} & \textbf{70.45} & \textbf{78.28} & \textbf{76.71} & \underline{80.17} & \textbf{65.45} \\
Gemma4 + Ours, layer 25
& \underline{99.03} & \textbf{95.10} & 34.43 & \underline{66.82} & 76.03 & 65.07 & 79.66 & \underline{63.10} \\
\bottomrule
\end{tabular}
}
\caption{
Ablation on world-model feature extraction layers.
Layer 24 achieves the best overall performance, suggesting that intermediate-to-late world-model representations provide the most transferable geometry-aware supervision.
}
\label{tab:layer_ablation}
\end{table*}

\paragraph{Denoising step selection.}
Table~\ref{tab:step_ablation} evaluates the number of denoising steps used when extracting teacher features.
Using zero or one denoising step improves over the base model, but both variants plateau at an overall score of $63.00$.
Using two denoising steps achieves the best overall performance of $65.45$, while using three steps reduces the overall score to $60.23$.
This indicates that the teacher signal is most useful after a small amount of spatial prediction, but excessive denoising may make the representation too generation-specific and less suitable for aligning the frozen VLM visual interface.

\begin{table*}[t]
\centering
\small
\setlength{\tabcolsep}{4.5pt}
\resizebox{\textwidth}{!}{
\begin{tabular}{lcccccccc}
\toprule
\textbf{Model}
& \textbf{Ctrl-A}
& \textbf{Ctrl-B}
& \textbf{COCO-1}
& \textbf{COCO-2}
& \textbf{VG-1}
& \textbf{VG-2}
& \textbf{VSR}
& \textbf{Overall} \\
\midrule
Gemma4
& 94.17 & 79.90 & \textbf{40.57} & 64.55 & 69.14 & 68.49 & 74.36 & 61.70 \\
Gemma4 + Ours, 0 step
& 97.57 & 93.14 & 34.88 & 66.36 & 75.34 & \underline{71.23} & 78.80 & \underline{63.00} \\
Gemma4 + Ours, 1 step
& 98.06 & \textbf{95.10} & 34.43 & \underline{66.82} & \underline{76.90} & 65.07 & 78.63 & \underline{63.00} \\
Gemma4 + Ours, 2 steps
& \textbf{99.51} & \underline{93.14} & \underline{37.46} & \textbf{70.45} & \textbf{78.28} & \textbf{76.71} & \textbf{80.17} & \textbf{65.45} \\
Gemma4 + Ours, 3 steps
& 95.63 & 88.73 & 34.25 & 68.18 & 70.00 & \underline{71.23} & 72.31 & 60.23 \\
\bottomrule
\end{tabular}
}
\caption{
Ablation on denoising step selection.
Using two denoising steps gives the best overall performance, while using three steps leads to clear degradation.
}
\label{tab:step_ablation}
\end{table*}

\paragraph{Additional hyperparameter ablation.}
We additionally ablate the alignment loss coefficient $\lambda_{\mathrm{align}}$ in Appendix~\ref{app:lambda_ablation}.
The best performance is obtained with $\lambda_{\mathrm{align}}=0.10$, suggesting that world-model alignment should be strong enough to inject useful spatial structure into the visual pathway, but not so strong that it disrupts the original VLM visual interface and its pretrained multimodal representation.

\section{Conclusion}
We presented \method{}, a vision-side distillation framework that transfers geometric structure from frozen camera-conditioned world models into VLMs. By fine-tuning only the vision encoder and multimodal projector, \method{} aligns visual tokens with world-model representations while keeping the language model frozen and requiring no world-model inference at test time.
Across spatial reasoning benchmarks and two VLM backbones, \method{} consistently improves the performance, especially on geometry-sensitive relations such as \textit{above}, \textit{under}, \textit{close}, and \textit{far}. Ablations show that prompt- and camera-aware intermediate world-model features are most effective.
Overall, our results suggest that improving the geometry encoded in visual tokens can mitigate spatial failures in VLMs without retraining the language model, highlighting world models as reusable teachers for spatially grounded multimodal intelligence.

\newpage
\bibliographystyle{abbrv}
{\small
\bibliography{egbib}}

\newpage
\appendix
\appendix
\section{More Details about LingBot-World-Fast}
\label{app:lingbot_world_fast}

GeoWorld-VLM uses LingBot-World-Fast as the default world-model teacher. 
LingBot-World-Fast is the efficient variant of LingBot-World, an open-source video-based world simulator designed to move beyond passive text-to-video generation toward interactive world modeling~\citep{team2026advancing}. 
Unlike conventional video generators that mainly synthesize short visual clips, LingBot-World is trained to model temporally extended visual dynamics, action-conditioned scene evolution, and long-range spatial consistency. 
These properties make it particularly suitable for our setting, where the goal is not to use the generated video as an additional test-time input, but to extract intermediate representations that encode how a static scene may evolve under controlled camera motion.

LingBot-World is developed through a multi-stage training pipeline. 
The first stage builds on a strong image-to-video diffusion prior, which provides high-fidelity visual generation and general spatiotemporal coherence. 
The middle-training stage further injects world knowledge and action controllability, enabling the model to maintain scene-level consistency over extended horizons and to respond to user-specified controls. 
The final post-training stage adapts the model for causal and efficient interactive generation through architectural adaptation and few-step distillation. 
LingBot-World-Fast corresponds to this efficient post-trained variant, which is optimized for low-latency rollout while preserving the structural and physical regularities learned by the larger world model.

A key feature of LingBot-World-Fast is its action-conditioned architecture. 
The model takes an initial image or video, noisy video latents, textual conditions, and user-defined action signals as inputs. 
Its action interface supports both continuous camera motion and discrete control signals. 
In the original LingBot-World formulation, camera motion is represented with geometric embeddings, while discrete actions such as keyboard-style controls are encoded as action tokens and injected into the diffusion transformer through adaptive normalization. 
As a result, the hidden states of the model are shaped not only by image appearance and text semantics, but also by the implied camera trajectory and the corresponding counterfactual scene transformation.

This design is aligned with the motivation of GeoWorld-VLM. 
Spatial relations such as \textit{behind}, \textit{under}, \textit{between}, \textit{close}, and \textit{far} are often difficult to infer from object identity alone. 
They depend on viewpoint, occlusion, support, relative depth, and how object projections change under camera motion. 
Because LingBot-World-Fast is trained to predict visually coherent future states under controllable motion, its intermediate representations are expected to contain motion-aware geometric cues that are complementary to static visual encoders such as DINO or CLIP. 
This is why we use LingBot-World-Fast as a feature-level teacher rather than as a video generator during inference.

In our implementation, LingBot-World-Fast is kept frozen throughout training. 
For each input image, we construct an image-to-video teacher input using a lightweight text condition and a sampled egocentric camera trajectory. 
We then run the world model for a small number of denoising steps and extract an intermediate hidden state from the diffusion transformer as the teacher representation. 
The extracted representation is projected into a shared alignment space and used to supervise the VLM post-projector visual tokens. 
Importantly, the generated frames themselves are not used as additional inputs to the student VLM at test time. 
The world model is only used during training to provide geometry-aware supervision, so GeoWorld-VLM retains the same inference interface and computational cost as the adapted VLM backbone.

We select LingBot-World-Fast rather than the full LingBot-World-Base model for practical reasons. 
The fast variant offers a better trade-off between teacher quality and feature extraction cost, which is important because teacher representations must be extracted for thousands of training examples. 
At the same time, the model remains grounded in the same world-modeling pipeline as the larger base model, including long-horizon consistency, action conditioning, and causal interactive generation. 
Therefore, LingBot-World-Fast provides a computationally feasible source of camera-conditioned spatial supervision while preserving the core geometric properties needed by GeoWorld-VLM.

\section{Training Details}
\label{app:training_details}

This appendix provides additional implementation and training details for GeoWorld-VLM and the compared distillation baselines.

\paragraph{Training objective.}
For all GeoWorld-VLM experiments, the language model backbone is frozen, and we update only the vision encoder, the multimodal projector, and the lightweight alignment heads.
The training objective consists of three terms:
\begin{equation}
\mathcal{L}
=
\mathcal{L}_{\mathrm{task}}
+
\lambda_{\mathrm{align}}\mathcal{L}_{\mathrm{align}}
+
\lambda_{\mathrm{preserve}}\mathcal{L}_{\mathrm{preserve}} .
\end{equation}
Here, $\mathcal{L}_{\mathrm{task}}$ is the cross-entropy loss over multiple-choice option-letter logits.
$\mathcal{L}_{\mathrm{align}}$ is a cosine alignment loss between the projected student visual representation and the projected teacher representation.
$\mathcal{L}_{\mathrm{preserve}}$ is a normalized mean-squared error loss between the current student visual representation and that of the frozen original VLM, which is used to reduce visual-interface drift.
Unless otherwise specified, we set $\lambda_{\mathrm{align}}=0.10$ and $\lambda_{\mathrm{preserve}}=0.05$.

\paragraph{Optimization hyperparameters.}
All methods are trained for three epochs with AdamW.
We use a learning rate of $2\times10^{-5}$, weight decay of $0.01$, gradient accumulation of $1$, maximum gradient norm of $1.0$, and random seed $42$.
For Gemma4-based experiments, we use a batch size of $4$ for both training and evaluation.
For InternVL3.5-2B-based experiments, we generally use a batch size of $2$ for the main-suite experiments due to the larger visual-token cost introduced by dynamic image tiling.

\paragraph{Alignment heads.}
For teacher-student feature alignment, both the student visual features and the teacher features are passed through separate two-layer MLP projection heads.
The hidden dimension of the projection head is set to $1024$, and the final alignment dimension is set to $512$.
The alignment loss is computed after both student and teacher features are projected into this shared feature space.
This design avoids requiring the raw teacher and student hidden states to have the same dimensionality.

\paragraph{World-model teacher.}
For GeoWorld-VLM, we use a frozen camera-conditioned video world model as the teacher.
The main experiments use the LingBot-Fast image-to-video teacher with $9$ frames, $2$ teacher denoising steps, and camera perturbation enabled.
We extract the teacher representation from transformer block $24$.
The teacher prompt is:
\begin{quote}
\small
\textit{A slight camera motion with stable object layout and unchanged spatial relations.}
\end{quote}
This prompt is designed to induce mild viewpoint changes while preserving the object layout and spatial relations in the original image.

\paragraph{World-model visual configuration.}
For the LingBot/Wan-based teacher, images are processed at resolution $480 \times 832$ with $9$ video frames.
The target timestep is set to $300$, and the shift parameter is set to $5.0$.
The VAE stride is $(4,8,8)$ and the patch size is $(1,2,2)$.
The teacher transformer has hidden dimension $5120$ and $40$ transformer layers.
For the LingBot-Fast denoising schedule used in our main experiments, the selected timesteps are
\[
[999,\;957,\;899,\;702],
\]
with corresponding sigmas
\[
[0.9998,\;0.9580,\;0.8994,\;0.7024].
\]
Unless otherwise specified, we use the two-step teacher setting in the main reported results.

\paragraph{Gemma4 visual processing.}
For Gemma4, the vision patch size is $16$.
In our implementation, the observed visual input representation can be organized as patch/token features, and feature alignment is performed over visual token features after projecting both the student and teacher features to the shared $512$-dimensional alignment space.
This matches the design choice in GeoWorld-VLM: the supervision is applied at the VLM visual interface before the frozen language model consumes the visual tokens.

\paragraph{InternVL3.5 visual processing.}
For InternVL3.5-2B, we use its dynamic image tiling strategy.
Each tile has resolution $448 \times 448$.
In the main reported setting, the maximum number of visual tiles is set to $4$.
The preprocessing procedure selects the closest aspect-ratio tiling layout under the tile budget, resizes and crops the image into $448$-sized tiles, and adds a thumbnail when multiple tiles are used.
The prompt is constructed by inserting repeated \texttt{<IMG\_CONTEXT>} tokens according to the number of image tokens and the number of visual tiles.
The student visual features are then projected to the shared $512$-dimensional alignment space before computing the alignment loss.

\paragraph{DINOv3 distillation baseline.}
For the DINO baseline, we replace the camera-conditioned world-model teacher with a frozen DINOv3 ViT-S/16 encoder.
Images are resized to $224 \times 224$, yielding an approximate $14 \times 14$ spatial grid with patch size $16$.
When the Hugging Face processor is unavailable, we use a fallback preprocessing pipeline consisting of resizing to $224 \times 224$ followed by ImageNet normalization.
The DINO visual features are treated as static teacher features and are projected into the same $512$-dimensional alignment space as the student features.
The alignment loss is again computed using cosine distance.
For both Gemma4 and InternVL3.5-2B, the DINO baseline uses the same default loss weights, namely $\lambda_{\mathrm{align}}=0.10$ and $\lambda_{\mathrm{preserve}}=0.05$, and is trained for three epochs under the same downstream supervision.

\paragraph{Model-specific settings.}
For Gemma4 with the world-model teacher, the main experiments use the LingBot-Fast image-to-video teacher with $9$ frames, $2$ denoising steps, transformer block $24$, camera perturbation, and the layout-preserving prompt described above.
For InternVL3.5-2B with the world-model teacher, we use the same teacher configuration, while adapting the student-side preprocessing to InternVL's dynamic tiling mechanism.
For Gemma4 with DINOv3 and InternVL3.5-2B with DINOv3, the only change is replacing the world-model representation with DINOv3 visual features; the optimization hyperparameters and loss weights are kept the same for a controlled comparison. The time consumed to use our method on Gemma4 is about 5.5 hours if the training set is What'sup and VSR and the compute source is 2 Nvidia H200.

\section{Additional Ablation Analysis}
\label{app:ablation_details}

The main paper reports the ablation results on world-model layer selection and denoising step selection in Section~\ref{subsec:ablation_studies}.
Here, we provide additional analysis of these two design choices and further report the ablation on the alignment loss coefficient.

\subsection{Analysis of World-Model Layer Selection}
\label{app:layer_analysis}

The layer selection results in Table~\ref{tab:layer_ablation} show that layer 24 provides the strongest overall supervision among the tested layers.
This pattern suggests that the usefulness of world-model features is not monotonic with depth.
A slightly shallower layer may still retain more local appearance information and may not fully encode the camera- and prompt-conditioned spatial structure needed for downstream spatial reasoning.
A later layer, on the other hand, may become more specialized toward the teacher model's generative objective and less compatible with the frozen VLM visual interface.

Layer 24 appears to offer a better balance.
It has already integrated viewpoint-conditioned spatial cues, but its representation remains sufficiently general to serve as a transferable supervision signal.
This is consistent with the design goal of GeoWorld-VLM: the teacher feature should provide geometry-aware information without forcing the student visual tokens to imitate generation-specific representations too strongly.

\subsection{Analysis of Denoising Step Selection}
\label{app:step_analysis}

The denoising step results in Table~\ref{tab:step_ablation} show that teacher features are most effective when extracted after two denoising steps.
This suggests that a small amount of world-model prediction helps reveal the geometry induced by the sampled camera trajectory.
With zero or one denoising step, the teacher representation may still be close to the initial image-conditioned latent and may not fully expose viewpoint-dependent structure.
This explains why both settings improve over the base model but remain limited at an overall score of $63.00$.

However, more denoising is not necessarily better.
Using three denoising steps substantially reduces the overall score.
One possible explanation is that later denoising states become increasingly tied to video synthesis rather than representation transfer.
They may contain generation-specific artifacts or drift away from the original image evidence, making them less suitable for supervising the VLM visual pathway.
Therefore, GeoWorld-VLM benefits most from an intermediate predictive representation: it should be spatially enriched by the world model, but still anchored to the input image.

\subsection{Alignment Loss Coefficient}
\label{app:lambda_ablation}

\begin{table*}[h]
\centering
\small
\setlength{\tabcolsep}{4.5pt}
\resizebox{\textwidth}{!}{
\begin{tabular}{lcccccccc}
\toprule
\textbf{Model}
& \textbf{Ctrl-A}
& \textbf{Ctrl-B}
& \textbf{COCO-1}
& \textbf{COCO-2}
& \textbf{VG-1}
& \textbf{VG-2}
& \textbf{VSR}
& \textbf{Overall} \\
\midrule
Gemma4
& 94.17 & 79.90 & \textbf{40.57} & 64.55 & 69.14 & 68.49 & 74.36 & 61.70 \\
Gemma4 + Ours, $\lambda_{\mathrm{align}}=0.05$
& \underline{99.03} & \underline{92.65} & 36.21 & \underline{66.82} & 76.03 & 69.86 & 77.95 & 63.49 \\
Gemma4 + Ours, $\lambda_{\mathrm{align}}=0.10$
& \textbf{99.51} & \textbf{93.14} & \underline{37.46} & \textbf{70.45} & \textbf{78.28} & \textbf{76.71} & \textbf{80.17} & \textbf{65.45} \\
Gemma4 + Ours, $\lambda_{\mathrm{align}}=0.15$
& \underline{99.03} & 92.16 & 35.85 & 66.36 & \underline{76.90} & \underline{72.60} & \underline{79.49} & \underline{63.88} \\
Gemma4 + Ours, $\lambda_{\mathrm{align}}=0.20$
& 98.54 & 89.22 & 33.99 & 69.55 & 75.69 & 65.75 & \underline{79.49} & 62.64 \\
\bottomrule
\end{tabular}
}
\caption{
Ablation on the alignment loss coefficient.
$\lambda_{\mathrm{align}}=0.10$ achieves the best overall result, suggesting that world-model alignment should be strong enough to inject spatial structure but not so strong that it disrupts the original VLM visual interface.
The best result is shown in bold and the second-best result is underlined.
}
\label{tab:app_lambda_ablation}
\end{table*}

Table~\ref{tab:app_lambda_ablation} studies the effect of the alignment loss coefficient.
The best performance is obtained with $\lambda_{\mathrm{align}}=0.10$, which reaches an overall score of $65.45$.
When the coefficient is too small, the teacher signal may be insufficient to inject useful spatial structure into the student visual representation.
When the coefficient is too large, the adapted visual tokens may overfit to the world-model feature space and become less compatible with the frozen language model.
Therefore, the alignment loss is best understood as a controlled geometric regularizer rather than a replacement objective for VLM training.

\subsection{Temporal Feature Design of the World-Model Teacher}
\label{app:frame_feature_ablation}

\begin{table*}[h]
\centering
\small
\setlength{\tabcolsep}{4.5pt}
\resizebox{\textwidth}{!}{
\begin{tabular}{lcccccccc}
\toprule
\textbf{Model}
& \textbf{Ctrl-A}
& \textbf{Ctrl-B}
& \textbf{COCO-1}
& \textbf{COCO-2}
& \textbf{VG-1}
& \textbf{VG-2}
& \textbf{VSR}
& \textbf{Overall} \\
\midrule
Gemma4 + Ours, 5 frames
& 98.54 & \textbf{97.06} & 34.25 & 67.27 & 73.97 & \underline{71.23} & 78.12 & 62.77 \\
Gemma4 + Ours, 13 frames
& \textbf{99.51} & \underline{96.57} & 33.54 & 64.09 & 72.76 & 67.12 & 77.95 & 61.86 \\
Gemma4 + Ours, first-frame feature
& 98.54 & 95.10 & 32.74 & 67.73 & 73.10 & 65.75 & 79.32 & 61.92 \\
Gemma4 + Ours, last-frame feature
& 96.60 & 91.67 & \underline{34.88} & \underline{68.18} & \underline{76.55} & 66.44 & \textbf{80.51} & \underline{63.30} \\
Gemma4 + Ours, 9 frames + mean pooling
& \textbf{99.51} & 93.14 & \textbf{37.46} & \textbf{70.45} & \textbf{78.28} & \textbf{76.71} & \underline{80.17} & \textbf{65.45} \\
\bottomrule
\end{tabular}
}
\caption{
Ablation on the temporal feature design of the world-model teacher.
The full method uses 9 generated frames and average pooling over all teacher features.
This configuration achieves the best overall performance, suggesting that useful spatial cues are distributed across the generated trajectory rather than being concentrated in a single frame.
The best result is shown in bold and the second-best result is underlined.
}
\label{tab:app_frame_feature_ablation}
\end{table*}

Table~\ref{tab:app_frame_feature_ablation} studies how the temporal design of the world-model teacher affects spatial reasoning performance on What’sUp+VSR.
Using 9 frames with mean-pooled teacher features achieves the best overall score of $65.45$.
Reducing the number of frames to 5 leads to a lower overall score, suggesting that a shorter generated trajectory may provide insufficient spatial dynamics.
Increasing the number of frames to 13 also decreases performance, which indicates that a longer trajectory may introduce additional visual or temporal noise into the teacher signal.

We further compare mean pooling with single-frame feature extraction.
Using only the first-frame feature obtains an overall score of $61.92$, while using only the last-frame feature reaches $63.30$.
Both are worse than mean pooling across all frames.
This result suggests that the teacher signal is not fully captured by any individual generated frame.
Instead, aggregating features across the generated trajectory provides a more stable and informative spatial representation.
Therefore, we use 9 frames with mean-pooled teacher features as the default configuration in our main experiments.

\subsection{Teacher Choice and Preservation Objective}
\label{app:teacher_preserve_ablation}

\begin{table*}[h]
\centering
\small
\setlength{\tabcolsep}{4.5pt}
\resizebox{\textwidth}{!}{
\begin{tabular}{lcccccccc}
\toprule
\textbf{Model}
& \textbf{Ctrl-A}
& \textbf{Ctrl-B}
& \textbf{COCO-1}
& \textbf{COCO-2}
& \textbf{VG-1}
& \textbf{VG-2}
& \textbf{VSR}
& \textbf{Overall} \\
\midrule
Gemma4 + Ours, w/o preservation
& \underline{99.03} & \textbf{94.61} & \underline{35.32} & 68.18 & 74.66 & 67.12 & 78.97 & 63.20 \\
Gemma4 + VGGT teacher
& 98.06 & 93.14 & 34.61 & \underline{70.00} & \underline{77.07} & \underline{69.86} & \underline{79.83} & \underline{63.65} \\
Gemma4 + Ours
& \textbf{99.51} & \underline{93.14} & \textbf{37.46} & \textbf{70.45} & \textbf{78.28} & \textbf{76.71} & \textbf{80.17} & \textbf{65.45} \\
\bottomrule
\end{tabular}
}
\caption{
Ablation on the teacher choice and the preservation objective.
Removing the preservation loss degrades overall performance, indicating that preserving the original visual interface is important when aligning the student representation to world-model features.
Replacing the image-to-video world-model teacher with VGGT also reduces the overall score, suggesting that dynamic world-model features provide more effective spatial supervision than static geometric features.
The best result is shown in bold and the second-best result is underlined.
}
\label{tab:app_teacher_preserve_ablation}
\end{table*}

Table~\ref{tab:app_teacher_preserve_ablation} evaluates two additional design choices.
First, removing the preservation loss reduces the overall score from $65.45$ to $63.20$.
This indicates that direct alignment to the world-model feature space may disrupt the original VLM visual interface if no constraint is used to preserve the pretrained representation.
The preservation loss therefore serves as an important regularizer, allowing the model to absorb geometry-aware teacher information while remaining compatible with the frozen language model.

Second, replacing the image-to-video world-model teacher with VGGT leads to an overall score of $63.65$, which is lower than the full method.
Although VGGT provides strong static geometric representations, it does not explicitly model how the scene evolves under camera-conditioned generation.
In contrast, the world-model teacher exposes the student to spatial cues along a generated visual trajectory.
This comparison supports our hypothesis that dynamic world-model features are more suitable for improving spatial relation understanding in VLMs.

\section{SAT Results by Ego-Motion Split}
\label{app:sat_ego_split}

\begin{table*}[h]
\centering
\small
\setlength{\tabcolsep}{4.5pt}
\resizebox{\textwidth}{!}{
\begin{tabular}{llcccccccc}
\toprule
\textbf{Split}
& \textbf{Model}
& \textbf{ActCons}
& \textbf{EgoM}
& \textbf{GoalAim}
& \textbf{ObjectM}
& \textbf{Perspect}
& \textbf{Val Acc.}
& \textbf{Test Acc.}
& \textbf{Overall} \\
\midrule
\multirow{3}{*}{EgoM only}
& Gemma4 + Ours
& -- & 52.43 & -- & -- & -- & 51.23 & 60.87 & 52.43 \\
& Gemma4 + FT-only
& -- & \underline{56.22} & -- & -- & -- & \underline{54.94} & \textbf{65.22} & \underline{56.22} \\
& Gemma4 + DINO
& -- & \textbf{63.78} & -- & -- & -- & \textbf{63.58} & \textbf{65.22} & \textbf{63.78} \\
\midrule
\multirow{3}{*}{Non-Ego}
& Gemma4 + Ours
& 45.55 & -- & \textbf{69.78} & \textbf{73.37} & \underline{57.46} & 58.11 & \textbf{59.06} & \textbf{58.24} \\
& Gemma4 + FT-only
& \textbf{46.36} & -- & \underline{68.13} & \underline{72.28} & \underline{57.46} & \textbf{58.47} & 55.12 & \underline{58.03} \\
& Gemma4 + DINO
& \underline{45.82} & -- & 63.74 & 71.20 & \textbf{60.09} & \underline{57.64} & \underline{55.91} & 57.41 \\
\bottomrule
\end{tabular}
}
\caption{
SAT results after separating ego-motion questions from non-ego questions.
The EgoM-only split shows that our world-model alignment performs worse than both FT-only and DINO on ego-motion reasoning.
On the non-Ego split, our method achieves the best overall accuracy, suggesting that world-model alignment is more effective for non-ego spatial reasoning than for ego-motion reasoning.
The best result within each split is shown in bold and the second-best result is underlined.
}
\label{tab:app_sat_ego_split}
\end{table*}

Table~\ref{tab:app_sat_ego_split} reports SAT results after separating ego-motion questions from the remaining question types.
This split is motivated by an important limitation of our current world-model alignment strategy.
Although the teacher is generated from a camera-conditioned image-to-video world model, our current implementation aggregates teacher features through mean pooling.
This design is effective for extracting stable spatial structure, but it may suppress temporal order and motion direction, both of which are essential for ego-motion reasoning.

The EgoM-only results confirm this limitation.
Our method obtains an EgoM score of $52.43$, which is lower than FT-only by $3.79$ points and lower than DINO by $11.35$ points.
This suggests that the current world-model representation is not sufficient for modeling ego-centric motion in SAT.
In particular, average pooling over generated frames may convert a temporally ordered motion signal into a static scene-level representation, making it difficult for the model to infer self-motion.

For the non-Ego split, our method achieves the best overall score of $58.24$, outperforming FT-only by $0.21$ points and DINO by $0.83$ points.
Although the margin over FT-only is modest, the result suggests that world-model alignment is beneficial once ego-motion questions are excluded.
The gains are mainly reflected in GoalAim and ObjectM, where our method achieves the best scores among the compared methods.
This indicates that the proposed alignment is more suitable for object-centric and goal-oriented spatial reasoning than for ego-motion reasoning.

Overall, the SAT analysis reveals a clear boundary of the proposed method.
World-model alignment improves spatial representation on What’sUp+VSR and achieves the best overall result on the non-Ego SAT split, but it struggles with ego-motion reasoning.
This observation motivates future extensions that preserve temporal directionality, such as temporal-aware pooling, order-sensitive teacher alignment, or explicit ego-motion supervision.
\section{Qualitative Case Studies}
\label{app:case_studies}
We provide qualitative examples from EmbSpatial-Bench to further illustrate how GeoWorld-VLM changes the spatial behavior of the base VLM.
As shown in Table~\ref{tab:case_studies_ours_only}, Cases 1--5 highlight examples where only our method predicts the correct answer, while the raw model, FT-only baseline, and DINO-aligned baseline all fail.
These cases suggest that task supervision or static visual feature alignment alone may still be insufficient for relations that require geometric structure, such as vertical placement, left-right ordering, and viewpoint-dependent distance judgment.
In contrast, GeoWorld-VLM benefits from camera-conditioned world-model supervision, which provides a more geometry-aware training signal for adapting the visual interface of the VLM.
Table~\ref{tab:case_studies_raw_wrong} further shows cases where our method corrects errors made by the raw model, while FT-only or DINO alignment may also succeed.
Together, these examples support our main finding that world-model alignment can improve spatial reasoning not merely by memorizing task labels, but by encouraging the visual pathway to preserve spatial cues that are useful before language reasoning begins.


\subsection{Cases Where Only Our Method Is Correct}

\begin{table*}[t]
\centering
\small
\setlength{\tabcolsep}{4pt}
\renewcommand{\arraystretch}{1.05}
\begin{tabular}{T{0.24\textwidth} T{0.34\textwidth} T{0.34\textwidth}}
\toprule
Image & Question / Ground Truth & Model Predictions \\
\midrule

\caseimg{case01.pdf}
&
\textbf{Q:} In the image, how do the positions of picture and ottoman interact with each other?

\textbf{GT:} The picture is above the ottoman.
&
Raw: The picture is on the left side of the ottoman. \xmark

FT-only: The picture is on the left side of the ottoman. \xmark

DINO: The picture is on the left side of the ottoman. \xmark

Ours: The picture is above the ottoman. \cmark
\\
\midrule

\caseimg{case02.pdf}
&
\textbf{Q:} In the image, how do the positions of television and bar interact with each other?

\textbf{GT:} The television is left of the bar.
&
Raw: The television is on the right side of the bar. \xmark

FT-only: The television is on the right side of the bar. \xmark

DINO: The television is on the right side of the bar. \xmark

Ours: The television is left of the bar. \cmark
\\
\midrule

\caseimg{case03.pdf}
&
\textbf{Q:} Among the listed objects, which one is farthest from your current location in the image?

\textbf{GT:} plant
&
Raw: bathtub \xmark

FT-only: bathtub \xmark

DINO: bathtub \xmark

Ours: plant \cmark
\\
\midrule

\caseimg{case04.pdf}
&
\textbf{Q:} From your viewpoint, which object among the options is the nearest within the image?

\textbf{GT:} chest
&
Raw: bed \xmark

FT-only: curtain \xmark

DINO: bed \xmark

Ours: chest \cmark
\\
\midrule

\caseimg{case05.pdf}
&
\textbf{Q:} What is the spatial relationship between chair and bicycle in the image?

\textbf{GT:} The chair is at the right side of the bicycle.
&
Raw: The chair is below the bicycle. \xmark

FT-only: The chair is below the bicycle. \xmark

DINO: The chair is below the bicycle. \xmark

Ours: The chair is at the right side of the bicycle. \cmark
\\

\bottomrule
\end{tabular}
\caption{Qualitative examples where only our LingBot-teacher alignment method predicts the correct spatial answer.}
\label{tab:case_studies_ours_only}
\end{table*}

\subsection{Cases Where Our Method Corrects the Raw Model}

\begin{table*}[t]
\centering
\small
\setlength{\tabcolsep}{4pt}
\renewcommand{\arraystretch}{1.05}
\begin{tabular}{T{0.24\textwidth} T{0.34\textwidth} T{0.34\textwidth}}
\toprule
Image & Question / Ground Truth & Model Predictions \\
\midrule

\caseimg{case06.pdf}
&
\textbf{Q:} Which object, in relation to your current position, holds the nearest placement in the image?

\textbf{GT:} pillow
&
Raw: pool table \xmark

FT-only: pillow \cmark

DINO: pillow \cmark

Ours: pillow \cmark
\\
\midrule

\caseimg{case07.pdf}
&
\textbf{Q:} What is the spatial configuration between monitor and lamp in relation to each other within the image?

\textbf{GT:} The monitor is beneath the lamp.
&
Raw: The monitor is on the left side of the lamp. \xmark

FT-only: The monitor is on the left side of the lamp. \xmark

DINO: The monitor is beneath the lamp. \cmark

Ours: The monitor is beneath the lamp. \cmark
\\
\midrule

\caseimg{case08.pdf}
&
\textbf{Q:} What is the spatial arrangement of wardrobe and curtain in the image concerning each other?

\textbf{GT:} The wardrobe is on the right side of the curtain.
&
Raw: The wardrobe is outside the curtain. \xmark

FT-only: The wardrobe is on the right side of the curtain. \cmark

DINO: The wardrobe is on the right side of the curtain. \cmark

Ours: The wardrobe is on the right side of the curtain. \cmark
\\
\midrule

\caseimg{case09.pdf}
&
\textbf{Q:} How are printer and paper positioned in relation to each other in the image?

\textbf{GT:} The printer is at the left side of the paper.
&
Raw: The printer is at the right side of the paper. \xmark

FT-only: The printer is at the left side of the paper. \cmark

DINO: The printer is at the left side of the paper. \cmark

Ours: The printer is at the left side of the paper. \cmark
\\
\midrule

\caseimg{case10.pdf}
&
\textbf{Q:} What is the spatial arrangement of picture and bin in the image concerning each other?

\textbf{GT:} The picture is above the bin.
&
Raw: The picture is left of the bin. \xmark

FT-only: The picture is above the bin. \cmark

DINO: The picture is above the bin. \cmark

Ours: The picture is above the bin. \cmark
\\

\bottomrule
\end{tabular}
\caption{Qualitative examples where our method corrects errors made by the raw model. FT-only or DINO alignment may also succeed on some of these examples, but the raw VLM fails.}
\vspace{-0.8cm}
\label{tab:case_studies_raw_wrong}
\end{table*}

\section{Limitations and Broader Impact.}
\label{app:limitations}

GeoWorld-VLM has several limitations. First, our method depends on the quality of the frozen world-model teacher: if the teacher encodes inaccurate viewpoint dynamics or weak scene geometry, the extracted features may provide noisy supervision. Second, although GeoWorld-VLM requires no world-model inference at test time, training still introduces additional cost because teacher features must be extracted from a large video world model. Finally, the method remains sensitive to feature extraction choices such as teacher layer and denoising step, and the gains are not uniform across all spatial relations or subsets. Future work could explore stronger and more efficient teachers, broader embodied evaluations, and more principled criteria for selecting geometry-aware teacher representations.
GeoWorld-VLM aims to improve the spatial reasoning ability of vision-language models by enhancing the visual representations consumed by a frozen language model. 
This capability could have positive impacts on embodied AI, assistive agents, robotics, navigation, and multimodal systems that need more reliable understanding of object layout and spatial relations. 
At the same time, stronger spatial perception may also increase the capability of systems used in surveillance, autonomous decision-making, or other safety-sensitive applications where incorrect spatial judgments could cause harm. 
Our work is a research study on benchmarked spatial reasoning rather than a deployed system, and we do not release new high-risk datasets, pretrained generative models, or autonomous agents. 
Future deployments of spatially enhanced VLMs should include careful evaluation under domain-specific safety requirements, robustness testing, and human oversight in high-stakes settings.
\newpage
\end{document}